\documentclass[sn-mathphys,Numbered]{sn-jnl}% Math and Physical Sciences Reference Style
\usepackage{graphicx}%
\usepackage{multirow}%
\usepackage{amsmath,amssymb,amsfonts}%
\usepackage{amsthm}%
\usepackage{mathrsfs}%
\usepackage[title]{appendix}%
\usepackage{xcolor}%
\usepackage{textcomp}%
\usepackage{manyfoot}%
\usepackage{booktabs}%
\usepackage{algorithm}%
\usepackage{algorithmicx}%
\usepackage{algpseudocode}%
\usepackage{listings}%
\usepackage{tabularx}%
\usepackage{color}%
\usepackage{soul}%
\usepackage[utf8]{inputenc} %unicode support
\begin{document}

\title{Static and multivariate-temporal attentive fusion transformer for readmission risk prediction}


\author[1,2]{\fnm{Zhe} \sur{Sun}}\email{sunzhe@gs.zzu.edu.cn}

\author*[2]{\fnm{Runzhi} \sur{Li}}\email{rzli@ha.edu.cn}

\author[1,2]{\fnm{Jing} \sur{Wang}}\email{wangjingzzu@gs.zzu.edu.cn}

\author[2]{\fnm{Gang} \sur{Chen}}\email{gchen@ha.edu.cn}

\author[3]{\fnm{Siyu} \sur{Yan}}\email{yansiyu@fuwaihospital.org}

\author[3]{\fnm{Lihong} \sur{Ma}}\email{mlh8168@163.com}

\affil*[1]{\orgdiv{School of Computer and Artificial Intelligence}, \orgname{Zhengzhou University}, \orgaddress{\city{Zhengzhou}, \postcode{450000}, \country{China}}}

\affil[2]{\orgdiv{Cooperative Innovation Center of Internet Healthcare}, \orgname{Zhengzhou University}, \orgaddress{\city{Zhengzhou}, \postcode{450000}, \country{China}}}

\affil[3]{\orgdiv{Fuwai Hospital, National Center for Cardiovascular Diseases}, \orgname{Chinese Academy of Medical Sciences and Peking Union Medical College}, \orgaddress{\city{Beijing}, \postcode{100000},\country{China}}}


%%================================%%
%% Sample for structured abstract %%
%%================================%%

 \abstract{\textbf{Background:} Accurate short-term readmission prediction of ICU patients is significant in improving the efficiency of resource assignment by assisting physicians in making discharge decisions. Clinically, both individual static static and multivariate temporal data collected from ICU monitors play critical roles in short-term readmission prediction. Informative static and multivariate temporal feature representation capturing and fusion present challenges for accurate readmission prediction. 
 \textbf{Methods:}We propose a novel static and multivariate-temporal attentive fusion transformer (SMTAFormer) to predict short-term readmission of ICU patients by fully leveraging the potential of demographic and dynamic temporal data. In SMTAFormer, we first apply an MLP network and a temporal transformer network to learn useful static and temporal feature representations, respectively. Then, the well-designed static and multivariate temporal feature fusion module is applied to fuse static and temporal feature representations by modeling intra-correlation among multivariate temporal features and  constructing inter-correlation between static and multivariate temporal features.
 \textbf{Results:} We construct a readmission risk assessment (RRA) dataset based on the MIMIC-III dataset. The extensive experiments show that SMTAFormer outperforms advanced methods, in which the accuracy of our proposed method is up to 86.6\%, and the area under the receiver operating characteristic curve (AUC) is up to 0.717.
 \textbf{Conclusion:}  Our proposed SMTAFormer can efficiently capture and fuse static and multivariate temporal feature representations. The results show that SMTAFormer significantly improves the short-term readmission prediction performance of ICU patients through comparisons to strong baselines.}



\maketitle

\section{Introduction}\label{sec1}
Patients of intensive care unit (ICU) with short-term readmission unexpectedly~\cite{b1,b2,b3,b4,b5}, may lead to a longer hospital stays \cite{b56} and a higher mortality risk \cite{b1, b6}. In particular, the short-term readmission rate is one of the most important metrics for ICU quality assessment, which is strongly related to the efficiency of ICU resource assignment.

Clinically, physicians make discharge decisions according to their experience, which is subjective and error-prone. Recent works have indicated that there was a high correlation relationship between the clinical data of patients and the readmission risk \cite{b57, b58}. With the development of artificial intelligence (AI) techniques, data-driven based decision methods can weaken the influence of subjective factors. A predictive model can predict the readmission risk of inpatients within short-term after discharge \cite{2018Machine}, conducing to medical resource assignment. Recently, a lot of AI methods have been applied to medical data analysis and clinical decision making, including multi-scale convolution, recurrent unit (GRU) \cite{b29}, long short-term memory (LSTM) \cite{b30}, bidirectional long short-term memory (Bi-LSTM) \cite{b32}. Here, we focus on the static and multivariate temporal data. Static data refer to demographic data of patients, such as sex, age, BMI and so on. They remain constant throughout the hospitalization. Multivariate temporal data refer to time sequence data from physiological monitoring, such as blood pressure, blood oxygen, heart rate, and so on. They may vary in real time. They contain different implications for indicating the status of patients. Thus, how to learn informative feature representations from static and multivariate-temporal data and then fuse them are two challenges for short-term readmission prediction.

\begin{figure} 
    \centering
        \includegraphics [width=0.95\textwidth]{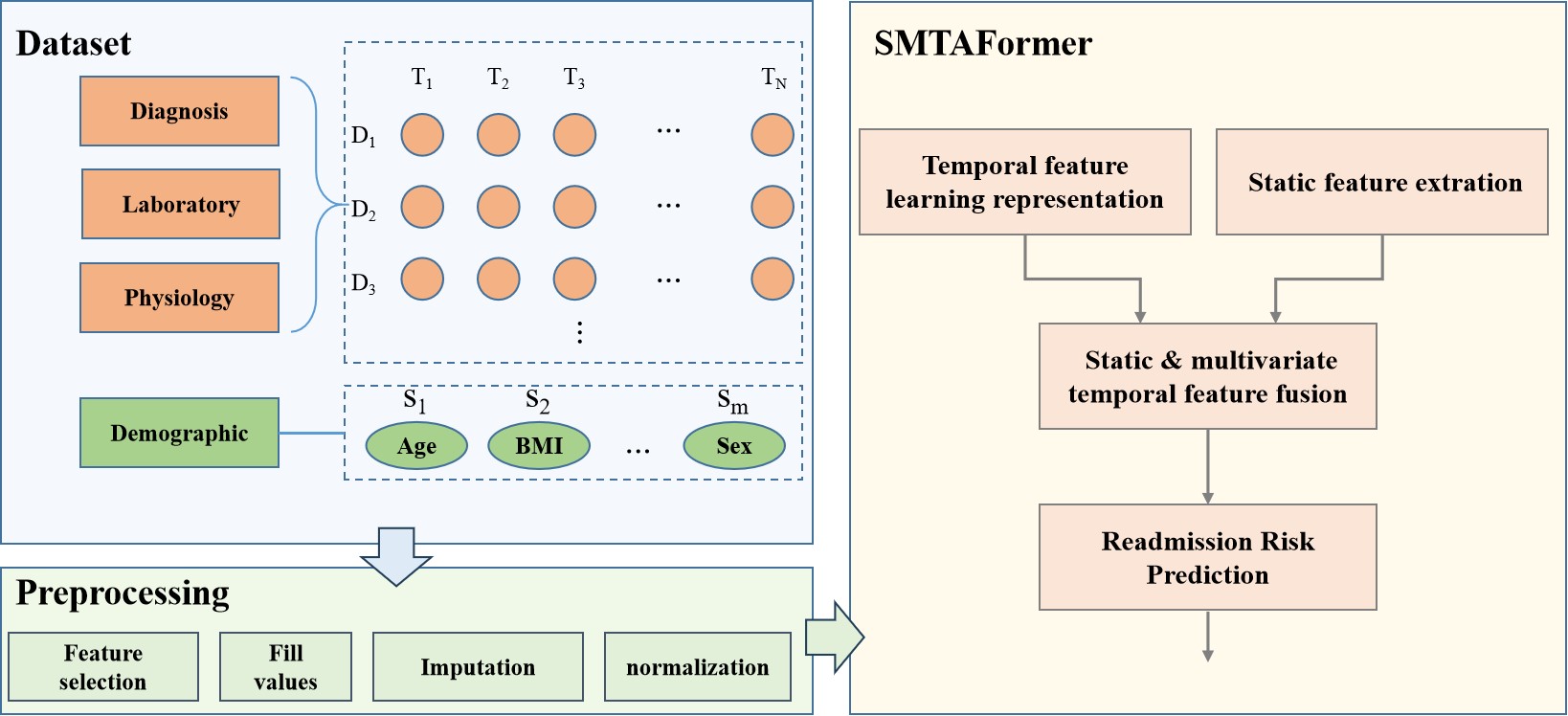}
    \caption{The flowchart of this main work. Static demographic data and diagnosis, laboratory, physiology monitoring data of patients are collected clinically for readmission risk prediction. Different preprocessing methods are taken to consitute datasets. We present SMTAFormer that extract static features and temporal features respectively, and then fuse static and multivariate temporal feature for readmission risk prediction.} 
    \label{fig.fc}
\end{figure}

To tackle these two challenges, we propose a static and multivariate-temporal attentive fusion network named SMTAFormer to predict the readmission risk of ICU patients. The flowchart is given in \ref{fig.fc}.

In SMTAFormer, we first apply an MLP network and a temporal transformer network to learn informative static and temporal feature representations, respectively. Next, a novel static and multivariate-temporal feature fusion module is proposed to fuse static and multivariate-temporal feature representations dynamically. Moreover, we construct a readmission risk assessment (RRA) dataset of ICU patients with essential hypertension based on the MIMIC-III dataset. The extensive experiments on the RRA dataset demonstrate the effectiveness of our method. The contributions of this paper are listed as follows:

\begin{itemize}
 \item We present a SMTAFormer to predict readmission risk of ICU patients by fully exploiting the static and multivariate-temporal data.

    \item We utilize an MLP network and a temporal transformer network to learn informative static and temporal feature representations accordingly, and then this work proposes a static and multivariate-temporal feature fusion module to integrate them adaptively. 

    \item We construct an RRA dataset for readmission risk prediction, which will be released soon. The results show that we significantly improves the readmission risk prediction performance through comparisons to strong baselines.
\end{itemize}

\section*{Related Work}label{RW}
In the past years, massive machine learning methods have been proposed to address readmission risk prediction tasks. Lin \cite{b9} used logistic regression (LR), support vector machine (SVM), random forest (RF) and naive Bayes (NB) to predict readmission risk of ICU patients. They used shallow extraction (e.g., slope and intercept) and deep extraction (e.g., quadratic terms and standard deviation) methods to mine the temporal feature representation. However, the experimental results show that there is little difference between them, and even deep extraction methods slightly reduce model performance. 

Recently, deep learning methods have been widely used to deal with patient electronic health records (EHRs) due to their powerful feature representation learning capability. Morid et al. \cite{b10} used a convolutional neural network (CNN) to extract multivariate temporal features of ICU patients. But most deep learning methods belong to recurrent neural networks (RNN) and their variants, such as GRU, LSTM, and Bi-LSTM, as a basis to extract the temporal features, including Patient2vec \cite{b11}, DeepRisk \cite{b12}, RETAIN \cite{b13}, and Dipole \cite{b14}. Given the independence of different features, multichannel techniques were introduced to deal with individual temporal features \cite{b16}. With the advent of self-attention mechanism \cite{b17}, transformers have gradually been used to tackle multivariate temporal data and have achieved considerable results \cite{b18, b19, b20, b21}. However, these method ignore the significance of static data and multivariate temporal data, which contain various information of patients.

Static data and multivariate-temporal data are the two different type of clinical data that describe the status of the patients. They are the same valuable, however it is different to tackle and represent them in data processing. According to the reference that there are mainly four methods to fuse static data and multivariate-temporal data. 
The first is splicing. One is splicing before entering the model \cite{b9}. And the other is splicing in the middle of the model \cite{b22}. Always the splicing methods are less persuasive. The second is fusion by loading to RNN. Li \cite{b23} set static data as an initial value of the unit of LSTM to fuse with temporal sequences data. This method finished fusion between static and temporal features while it did not expose the correlation between them. It is limited by the amnesia of the RNN model and has poor expressiveness for long time series features. The third is fusion based on gate mechanism. Lim \cite{b24}  constructed a deep network with gate mechanism to filter temporal data by high relation with static features in timing dimension. However, it missed correlation among multivariate-temporal features. The Last is fusion by attention mechanisms. This method can reflect the correlation between different modality data. For example An \cite{b25} and Zhang \cite{b26} that they use attention mechanism to fuse static and temporal data. However, it can result in the redundancy of static data, which would result in weakening the performance of models. In this work, we focus on a novel method that fuse
static data and multivariate-temporal data, in which we hope to extract the inter correlation between them and further the intra multivariate-temporal data correlation.

\section*{Problem Formulation}\label{Task Definitin}
In this work, based on static data and multivariate temporal data, we construct a deep learning models to predict readmission risk within 30 days for ICU patients.

Firstly, as the input of the model: $ X = \left\{ S, T \right\}$, we build numeric vectors $ S \in \mathbb{R}^{m} $ from the static features of the patients, such as age and sex. The numeric vector $T = \left\{ {X_{1}, X_{2}, \ldots, X_{n}} \right\}$is built on the multivariate temporal features of the patients, and $n$ is the size of multivariate temporal features. $X_{i} = \left\{ {x_{1}^{i}, x_{2}^{i}, \ldots, x_{t}^{i}} \right\}$ for each $X_{i}$, and $t$ is time sequence, $x_{j}^{i} \in \mathbb{R}^{d_{i}}\left( j \in \lbrack 1, 2, \ldots, t\rbrack \right)$.

For the outputs: $\overset{-}{y} \in (0,1)$. $\overset{-}{y}$ represents the readmission probability for ICU patients in 30 days. y is the ground truth. It is binary, where 0 indicates that patients are not readmitted within 30 days after discharge. While 1, indicates that patients are readmitted within 30 days.

Our objective is to adjust the parameters to optimize the prediction model. It is described as following.
\begin{equation}
    f_{\theta^{*}} = {argmin}_{f_{\theta} \in U}\left\lbrack {l}\left( {f_{\theta}(X),y} \right) + \lambda p\left( f_{\theta} \right) \right\rbrack
\end{equation}
where $f_{\theta}(X)$ is the prediction model. $\theta^{*}$ indicates the optimal parameter set. $l$ is the loss function. $\lambda p$ is the regularized term of $\theta$, and $\lambda$ is a hyperparameter. $U$ represents the field of parameters $\theta$.
\begin{figure} 
    \centering
        \includegraphics [width=0.95\textwidth]{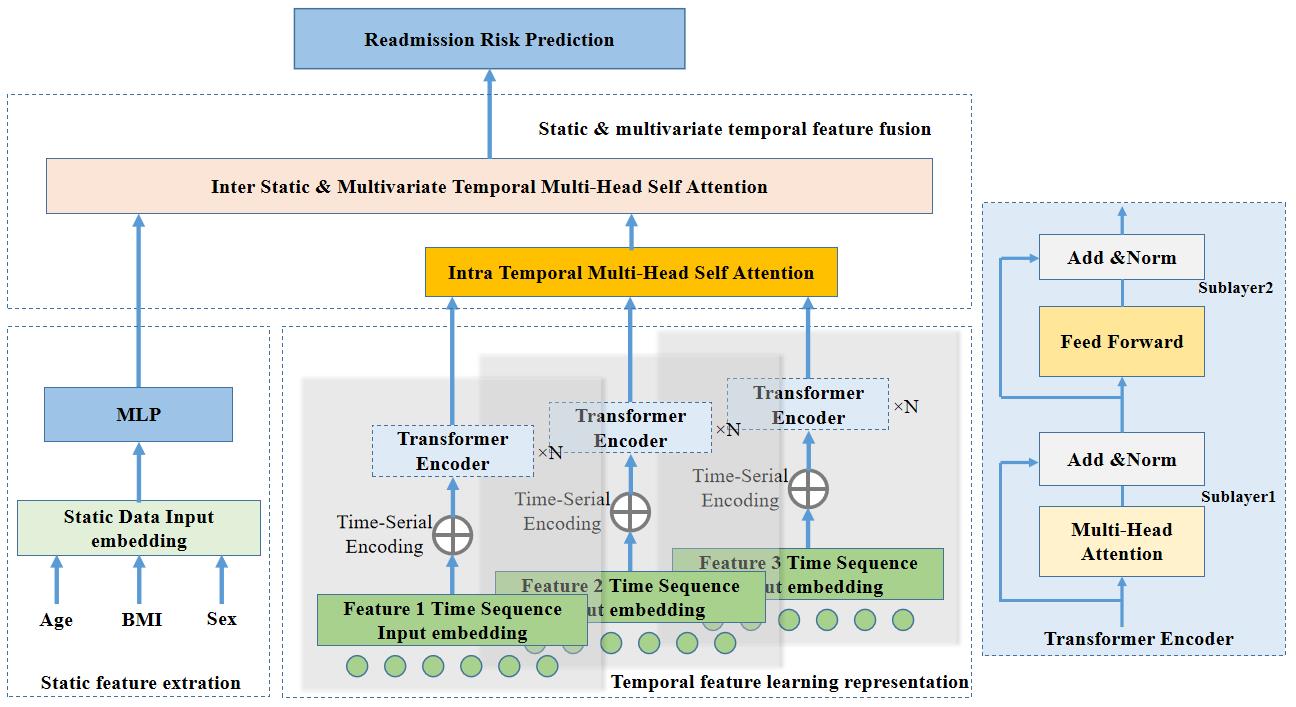}
    \caption{The architecture of SMTAFormer,in which the intra temporal multi-head self attention is used to extract correlations among multivariate temporal data and the inter static \& multivariate temporal multi-head self attention is designed for the fusion of static \& multivariate temporal data.} 
    \label{fig1}
\end{figure}

\section*{Method}\label{Methodology}
In this work, we propose SMTAFormer to predict readmission risk probability of inpatients with clinical data, including static and multivariate temporal features. SMTAFormer comprises static feature representation extraction, multivariate temporal feature representation extraction, static and multivariate temporal feature fusion, and readmission risk prediction.

First, we use a simple MLP network for static feature extraction. Then, we adopt the traditional transformer encoder for learning feature representation of dynamic temporal data. Next, we construct a attentive fusion network, in which an intra temporal multi-head self attention layer followed by an inter static and multivariate temporal multi-head self attention. Lastly, a fully connected layer is used to finish the readmission risk prediction. Figure \ref{fig1} shows the architecture of SMTAFormer.

\subsubsection*{Static feature representation extraction}

We construct a simple MLP network to extract static feature representations from static data, which is a combination of a fully-connected (FC) layer and and a ReLU activation function:
\begin{equation}
\overset {-}{S} = \mathrm{ReLU} \left({SW_{s}} + b_{s} \right),
\end{equation}
where $W_{s} \in \mathbb{R}^{d \times m}$ and $b_{s} \in \mathbb{R}^{d}$ are learnable parameters, and $\overset{-}{S} \in \mathbb{R}^{d}$ represents the generated static feature representations.

\subsubsection*{Multivariate temporal feature representation extraction}

For multivariate temporal data, we apply a temporal transformer network to capture informative multivariate temporal feature representations. First, we utilize a FC layer to encode the multivariate temporal data followed by positional encoding. Then we construct the multi-head self-attention layer, the feed forward layer and the Add \& Normalization layer for learning multivariate temporal feature representations. Finally, an average pooling layer is added to obtain useful multivariate temporal feature representations. It is described as follows:

Given multivariate temporal data $X_{i} \in \mathbb{R}^{t \times d_{i}}$. First, we adopt a linear transformation and a ReLU activation function to encode the output $X_{i}$.
\begin{equation}
    E^{i} = \mathrm{ReLU}\left( X_{i}W_{i}^{t} + b_{i}^{t} \right)
\end{equation}
where $W_{i}^{t} \in \mathbb{R}^{d \times d_{i}}$ and $b_{i}^{t} \in \mathbb{R}^{d}$ are learnable parameters. For the same temporal features of the different timing nodes, the learnable parameters are shared. $E^{i} = \left\{ e_{1}^{i},e_{2}^{i},\ldots,e_{t}^{i} \right\} \in \mathbb{R}^{t \times d}$. $E^{i}$ represents the time series after nonlinear transformation in a single channel. After that, we use positional encoding to add timing information to the time series.
\begin{equation}
    P^{i} = P + E^{i}
\end{equation}
Where $P \in \mathbb{R}^{t \times d}$ represents the position encoding matrix, $P^{i} \in \mathbb{R}^{t \times d}$ on behalf of the time series after adding timing information. Then $P^{i}$ is sent to the encoder to extract temporal features. Here, we introduce the encoder as follows.

Firstly, the multi-head self-attention mechanism is used to calculate the correlation among the time series.
\begin{equation}
    M_{i} = Concat\left( {{head}_{1},{head}_{2},\ldots,{head}_{h}} \right)W
\end{equation}
\begin{equation}
    {head}_{l} = Attention\left( {P^{i}W_{l}^{Q},P^{i}W_{l}^{K},P^{i}W_{l}^{V}} \right)
\end{equation}
\begin{equation}
    Attention\left( {Q,K,V} \right) = softmax\left( \frac{QK^{T}}{\sqrt{d}} \right)V
\end{equation}
where $W_{l}^{Q} \in \mathbb{R}^{d \times d}$, $W_{l}^{K} \in \mathbb{R}^{d \times d}$, $W_{l}^{V} \in \mathbb{R}^{d \times d}$, and $W \in \mathbb{R}^{h*d \times d}$ are training parameters. $W_{l}^{Q}$, $W_{l}^{K}$, and $W_{l}^{V}$ represent the query, Key, and Value matrix in the self-attention mechanism, respectively. The linear transformation matrix $W$ makes the output $M_{i}$ of the same shape as $P^{i}$. The corresponding elements are added by $M_{i}$ to the original input $P^{i}$ and through layer normalization.
\begin{equation}
    M_{i}^{'} = LayerNormalization\left( {M_{i} + P^{i}} \right)
\end{equation}
where $M_{i}^{'} \in \mathbb{R}^{t \times d}$. Then the output $M_{i}^{'}$ is fed into the feedforward neural network, which is composed of two linear layers, an activation function $\mathrm{ReLU}$, residual connection, and layer normalization.
\begin{equation}
    {\overset{\sim}{M}}_{i}^{'} = \mathrm{ReLU}\left( {M_{i}^{'}}W_{1} + b_{1} \right)W_{2} + b_{2}
\end{equation}
\begin{equation}
M_{i}^{''} = LayerNormalization\left( {M_{i}^{'} + {\overset{\sim}{M}}_{i}^{'}} \right)
\end{equation}
The final output $M^{final} \in \mathbb{R}^{t \times d}$ is obtained by stacking multiple encoders. Then we adopt an average pooling layer in the time dimension to obtain the temporal characteristics represented under the time length $t$.
\begin{equation}
    m_{i} = AveragePooling\left( M_{i}^{final} \right)
\end{equation}
where $m_{i} \in \mathbb{R}^{d}$.

In summary, we get the feature representations of the static and dynamic temporal data as $\overset{-}{S}$ and $\overset{-}{M} = \left\{ m_{1},m_{2},\ldots,m_{n} \right\}$ separately.

\subsection* {Static and multivariate temporal feature fusion}

In this work, we propose a inter static and multivariate temporal multi-head self-attention module to fuse static and multivariate temporal feature representations adaptively. First, it calculates the correlation among multivariate temporal feature representations. We get a weighted sum of them. Next, the embedded static feature representations is adopted as the source of query to fuse static and temporal features by another multi-head self-attention layer. It can avoid redundant computing operations caused by feeding static features to the model many times. Next is the explanation.

Firstly, $\overset{-}{M}$ denotes the correlation of single temporal data.
\begin{equation}
\overline{\overline{M}} = Concat\left( {{head}_{1}^{'},{head}_{2}^{'},...,{head}_{h}^{'}} \right)\overset{-}{W}
\end{equation}
\begin{equation}
    {head}_{l}^{'} = Attention\left( {\overset{-}{M}{W_{l}^{Q}}^{'},\overset{-}{M}{W_{l}^{K}}^{'},\overset{-}{M}{W_{l}^{V}}^{'}} \right)
\end{equation}
where ${W_{l}^{Q}}^{'} \in \mathbb{R}^{d \times d}$, ${W_{l}^{K}}^{'} \in \mathbb{R}^{d \times d}$, ${W_{l}^{V}}^{'} \in \mathbb{R}^{d \times d}$ and $\overset{-}{W} \in \mathbb{R}^{n*d \times d}$ are training parameters. $\overline{\overline{M}} \in \mathbb{R}^{n \times d}$ is the representation of characteristics with multi-head slef attention mechanism.

Next, a multi-head attention mechanism is used for the fusion of static and temporal features. We use the static features $\overset{-}{S}$ as the source of the query. and $\overset{-}{X}$ is the source of the key and value. 
\begin{equation}
    \overset{-}{X} = Concat\left( \overset{-}{S},\overset{-}{M} \right)
\end{equation}
\begin{equation}
    \overset{\sim}{X} = Concat\left( {{head}_{1}^{''},{head}_{2}^{''},...,{head}_{h}^{''}} \right)\overline{\overline{W}}
\end{equation}
\begin{equation}
    {head}_{l}^{''} = Attention\left( {\overset{-}{S}{W_{l}^{Q}}^{''},\overset{-}{X}{W_{l}^{K}}^{''},\overset{-}{X}{W_{l}^{V}}^{''}} \right)
\end{equation}
where ${W_{l}^{Q}}^{''} \in  \mathbb{R}^{d \times d}$, $
{W_{l}^{K}}^{''} \in  \mathbb{R}^{d \times d}$, ${W_{l}^{V}}^{''} \in  \mathbb{R}^{d \times d}$ and $\overline{\overline{W}} \in \mathbb{R}^{n*d \times d}$ are training parameters. $\overset{\sim}{X} \in  \mathbb{R}^{d}$ is a vector representation for the fusion of static and temporal features.

\subsection*{Readmission Task Prediction}
In the readmission task prediction module, we use two fully connected layers to perform the non-linear transformation. $\mathrm{ReLU}$ and $\mathrm{Sigmoid}$ are activation functions, respectively.  
\begin{equation}
    \overset{\sim}{y} = \mathrm{Sigmoid}\left( {\mathrm{ReLU}\left( {\overset{\sim}{X}W_{3} + b_{3}} \right)W_{4} + b_{4}} \right)
\end{equation}
where $W_{3} \in \mathbb{R}^{d \times r}$, $W_{4} \in   \mathbb{R}^{1 \times r}$, $b_{3} \in \mathbb{R}^r$ and $b_{4} \in   \mathbb{R}^{1}$ are learnable parameters. $\overset{\sim}{y}$ is the output of the model.

\section*{Experiments}\label{Experiments}
In this section, we introduce the RRA dataset, data preprocessing, experimental setup, and evaluation metrics. Then, the results and the correlation between static and multivariate temporal characteristics.

\subsection*{Dataset}
It should be emphasized that we choose patients in the ICU with essential hypertension as the readmission research object and construct the experimental dataset from MIMIC-III, named RRA. We follow the suggestions of local cooperative physicians. The benefit is to highlight the correlation between characteristics and disease and to improve model interpretation. 
 
 The MIMIC-III data set is an open-sourced medical dataset based on patient conditions in the MIT intensive care unit operated by MIT \cite{b8}. To build the RRA dataset of ICU patients with essential hypertension, we choose four types of static characteristics referenced by Lin \cite{b9}: age, sex, insurance, and race. They are vital signs for reflecting the condition of the patients. We selected 12 temporal characteristics consisting of heart rate, diastolic blood pressure, systolic blood pressure, mean blood pressure, respiration rate, oxygen saturation, body temperature, and Glasgow coma scale.
 
 Specifically, we select three types of International Classification of Diseases, Version 9 (ICD 9) codes associated with essential hypertension, including 4010, 4011, and 4019. Meanwhile, we remove three types of ICU records, one for patients younger than 18 years of age, the second is for patients who died in the ICU and the last for patients admitted by pregnancy. Based on this, we screen ICU records for ICU stays longer than 24 hours and less than 72 hours. Regarding the labels of the experimental dataset, we make the following definition: a patient returns to the ICU within 30 days after last leaving the ICU, or when the patient died within 30 days, we mark the label as 1, otherwise it is 0. 
 
 Finally, we get 10008 ICU records from ICU patients with essential hypertension, and 1110 data are readmitted to the ICU within 30 days. We describe the RRA dataset as shown in Table 1.

\begin{table*}[ht!]
\caption{Description of the Dataset\label{table1}}
\newcolumntype{C}{>{\centering\arraybackslash}X}
\resizebox{\textwidth}{!}{
\begin{tabular}{cccc}
\\ \hline
                                                              & \begin{tabular}[c]{@{}c@{}}Features of \\ different modes\end{tabular} & Description                                                                 & \begin{tabular}[c]{@{}c@{}}Number of \\ Features\end{tabular} \\ \hline
Features                                                      & \begin{tabular}[c]{@{}c@{}}Static \\ Features\end{tabular}             & Age, Gender, Insurance Type and Ethnicity                                                                                                                        & 4                                                             \\  
                                                              & \begin{tabular}[c]{@{}c@{}}Temporal \\ Features\end{tabular}           & \begin{tabular}[c]{@{}c@{}}Diastolic blood pressure, Glucose, Heart Rate, \\ Mean blood pressure, Oxygen saturation, \\ Respiratory rate, Systolic blood pressure, \\ Temperature, Glascow coma scale eye opening, \\ Glascow coma scale motor response, \\ Glascow coma scale verbal response \\ and Glascow coma scale total\end{tabular} & 12                                                            \\   
                                                              & Total                               &                                     & 16                                  \\ \hline
                                                              & Label   & Description                         & \begin{tabular}[c]{@{}c@{}}Number of \\ Labels\end{tabular}   
                                                              \\ \hline
Labels                                                        & 0                                                                      & Records of ICU readmission not occurred within 30 days                                                                                                                                                                            & 8898                                                          \\  
                                                              & 1                                                                      & Records of ICU readmission occurred within 30 days                           & 1110                               \\  
                                                              & Total                              &                                                                           & 10008                                                         \\ 
\bottomrule
\end{tabular}}
\end{table*} 
 
\subsection*{Data Preprocessing}
In the data preprocessing stage, the forward and backward filling method is used to interpolate the missing data. For continuous multivariate temporal data, average is as the unit value. For discrete multivariate-temporal data, we calculate the mode value as the interpolation. Figure 2 shows three examples of discrete multivariate temporal features for continuous and discrete multivariate temporal data, the z-score and one-hot encoding methods are used for standardization, respectively.

the training dataset and the test dataset are divided by 9: 1. We use 5-fold cross-validation and binary cross-entropy loss function in the training. Adam is used as the optimizer, with a learning rate of 0.001, the batch size is set to 32, and the epoch is set to 150. We adopt the early stop mechanism to prevent overfitting. 

\begin{figure}[ht!]
\centering
\includegraphics [width= 1\textwidth]{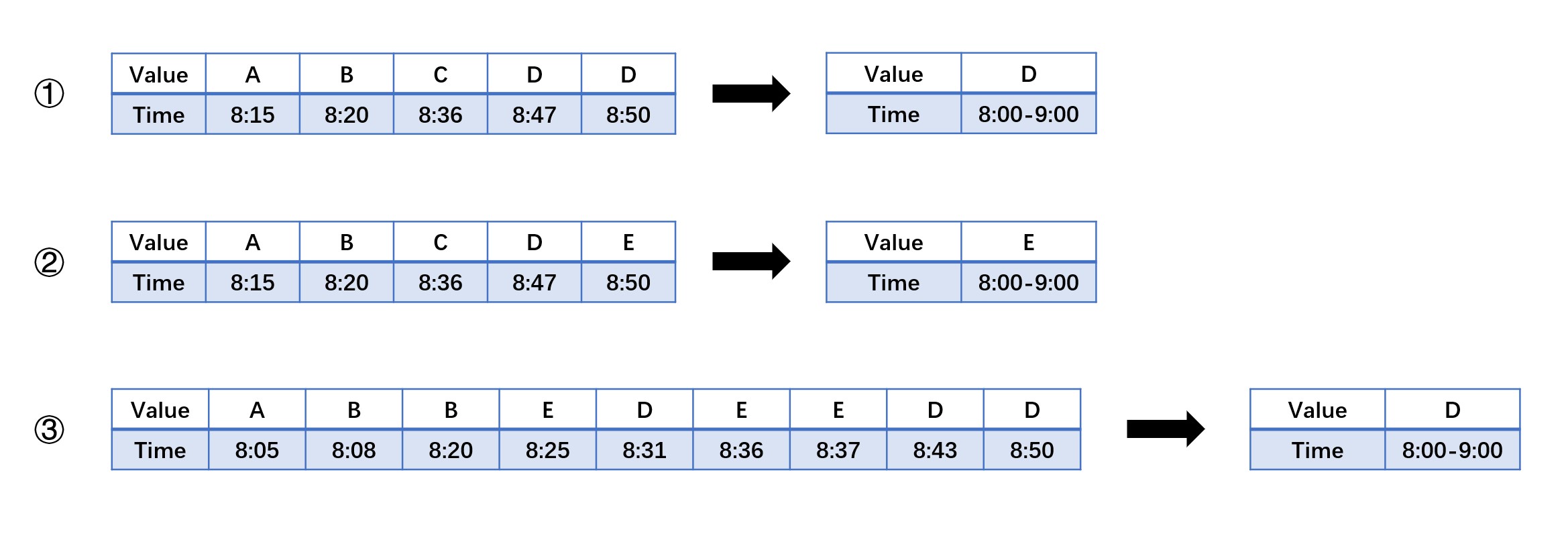}
\caption{Examples of selecting the discrete temporal feature within an hour.
\label{fig.2}}
\end{figure}

\subsection*{Contrast Methods}
 In the experiments, as the experimental contrast, we deploy three baseline methods. They are LR and LSTM and CNN+LSTM, in which all data are processed by single modality, we montage static data at each temporal step. To evaluate our proposed methods SMTAFormer, we make experiments by two stages. Firstly, we contrast different models for different types of data. Because there are three types of data in our dataset that consists of static features, continuous temporal features and discrete temporal features. In which we choose the classic Dense neural network and Multi-scale convolution network and Multi-scale convolution network with learning addition models for static features processing. LSTM and GRU are adopted for continuous temporal features analysis. For another modality, they are static data with temporal feature. We try to use CNN and LSTM models.

In the second stage, we focus on the fusion module. We compare three methods consist of SAF (Fusion of self-attention mechanism), CGRN  (Gating residual splicing network) and DSAF (Fusion of double self-attention mechanism). In addition, we use Transformer encoder for temporal feature representation. 
\subsection*{Evaluation Metrics}
In the experiments, we set metrics as follows: Accuracy Rate (ACC), Precision, recall, and the area under the ROC curve (AUC). 

\section{Results}\label{sec2}
We mainly analyze the experimental results from two views: feature extraction and feature fusion.
\begin{table}[ht!] 
\caption{Results on three baseline models for single modality processing.\label{table2}}
\newcolumntype{C}{>{\centering\arraybackslash}X}
\begin{tabularx}{\textwidth}{CCCCCCCCC}
\toprule
\multicolumn{3}{c}{\textbf{Methods}} & \textbf{ACC}            & \textbf{Precision}          & \textbf{Recall}          & \textbf{AUC} \\         
\midrule                                                                   
\multicolumn{3}{c}{LR}       & 0.690                & 0.533                 & 0.573                 & 0.603 \\          \multicolumn{3}{c}{LSTM}     & 0.743                & 0.563                  & 0.617                 & 0.638 \\         
\multicolumn{3}{c}{CNN+LSTM} & 0.767                & 0.575                 & 0.614                  & 0.650 \\         \bottomrule    
\end{tabularx}
\end{table}

\begin{table}[ht!] 
\caption{Results on multi-modality processing.\label{table3}}
\newcolumntype{C}{>{\centering\arraybackslash}X}
\begin{tabularx}{\textwidth}{CCCCCCCCC}
\toprule
\multicolumn{3}{c}{\textbf{Methods}} & \textbf{ACC}            & \textbf{Precision}          & \textbf{Recall}          & \textbf{AUC} \\   
\textbf{Static}    & \textbf{CT}        & \textbf{DT}        &                &                &                &   \\
\midrule                                                                   
DNN      & LSTM     & CNN    & 0.810                & 0.589                  & 0.630                 & 0.670    \\
MCN      & LSTM     & CNN    & 0.802                & 0.580                 & 0.618                  & 0.651    \\      
MCNL     & LSTM     & CNN    & \textbf{0.811}       & 0.591                 & \textbf{0.635}         & 0.668     \\
MCNL     & GRU     & CNN    & 0.807                & 0.587                  & 0.632                & 0.666        \\
DNN      & LSTM     & LSTM   & 0.799                & \textbf{0.592}       & \textbf{0.653}      & \textbf{0.682}  \\
\bottomrule
\end{tabularx}
\end{table}

Table 2 and Table 3 shows performance comparisons among the baseline models and our proposed multimodal and multichannel models. First, table 2 gives that CNN+LSTM[STC] achieves the best prediction results with 0.767 of accuracy and 0.650 of AUC among three baseline models. As for the multimodal and multichannel methods, table 3 lists that all the models outperform baseline models on four metrics. Particularly, the combination of DNN+LSTM+LSTM obtains the highest AUC value with 0.682. Base on the results of table 3, we choose DNN and LSTM for the static features and continuous temporal features processing in the following experiments. 

\begin{table}[ht!] 
\caption{Performance of Improved Models.\label{tab1e4}}
\newcolumntype{C}{>{\centering\arraybackslash}X}
\begin{tabularx}{\textwidth}{CCCCCCCCC}
\toprule
\multicolumn{3}{c}{\textbf{Methods}} & \textbf{ACC}            & \textbf{Precision}          & \textbf{Recall}          & \textbf{AUC} \\   
\textbf{Static}    & \textbf{Temporal}        & \textbf{Fusion}        &                &                &                &   \\
\midrule
DNN  & LSTM     & -        & 0.799           & 0.592          & 0.653          & 0.682         \\
DNN  & LSTM     & SAF       & 0.781          & 0.582          & 0.646          & 0.702          \\
DNN  & LSTM     & CGRN     & 0.825          & 0.597          & 0.633           & 0.673          \\
DNN  & LSTM     & CGRN+SA  & 0.815          & 0.600          & 0.653          & \textbf{0.704} \\
DNN  & LSTM+SA  & SAF       & 0.816         & 0.578          & 0.608         & 0.702          \\
DNN  & Encoder  & SAF       & 0.837         & 0.609          & 0.624         & 0.691          \\
DNN  & Encoder  & DSAF      & \textbf{0.866}  & \textbf{0.632} & \textbf{0.619}  & \textbf{0.702}          \\
\bottomrule
\end{tabularx}
\end{table}

\begin{table}[ht!] 
\caption{Performance Comparison of Stacking Different Number of Encoder Layers.\label{table5}}
\newcolumntype{C}{>{\centering\arraybackslash}X}
\begin{tabularx}{\textwidth}{CCCCCCC}
\toprule
\textbf{\begin{tabular}[c]{@{}c@{}}Numbers of \\ Encoder Layers\end{tabular}}	& \textbf{ACC}	& \textbf{Precision}
& \textbf{Recall}	& \textbf{AUC}\\
\midrule
1 & 0.836          & 0.598          & \textbf{0.619} & \textbf{0.717} \\
2 & \textbf{0.866} & \textbf{0.632} & \textbf{0.619}  & 0.702          \\
3 & 0.849          & 0.6          & 0.619         & 0.715         \\
\bottomrule
\end{tabularx}
\end{table}

In table 4, the comparative results of readmission risk prediction results with modal fusion are presented. Notably, the DSAF module stands out as it demonstrates superior performance, with an AUC reaches 0.702. Incorporating an attention mechanism across different time units within a single channel leads to enhancements in both ACC and AUC metrics for extracting temporal features. Particularly noteworthy is the Encoder method, showcasing the most significant improvement, with ACC and AUC reaching 0.837 and 0.691, respectively. Overall, the ensemble of DNN, encoder, and DSAF in SMTAFormer outperforms other comparative approaches. Moving to Table 5, it is observed that employing an encoder layer size of 2 yields optimal performance.

\subsection*{Visualization of the correlation between static and temporal features}
To ascertain the correlation between static and temporal features, samples labeled as 1are carefully curated and fed into the training model. Subsequently, the weight matrix of self-attention from various heads in the temporal and static fusion is derived. This weight matrix serves as the representation of the correlation coefficients between the static and temporal features, elegantly depicted in Figure 3.
\begin{figure}[ht!]
\centerline{
\includegraphics [width= 1\textwidth]{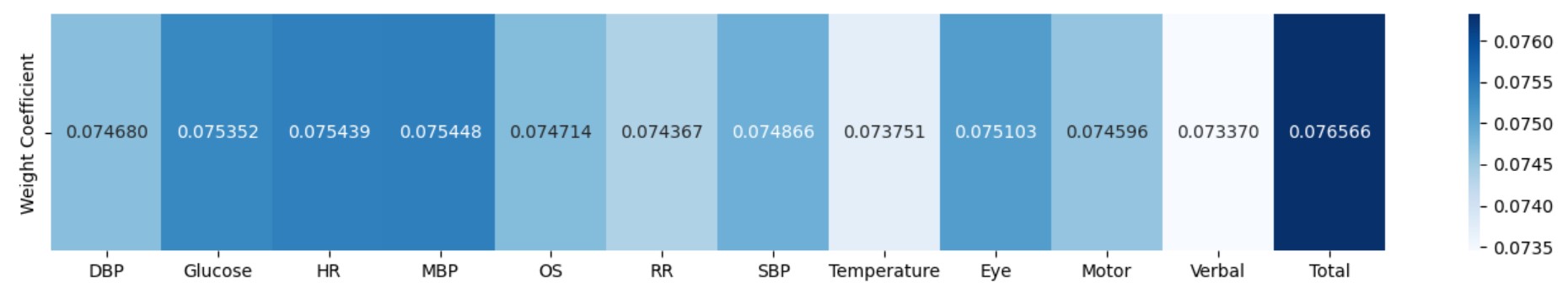}
}
\caption{Heat map of attention weight coefficients of 12 temporal features. DBP: Diastolic blood pressure, HR: Heart Rate, MBP: Mean blood pressure, OS: Oxygen saturation, RR: Respiratory rate, SBP: Systolic blood pressure, eye: Glasgow coma scale eye opening, Motor: Glasgow coma scale motor response, Verbal: Glasgow coma scale verbal response, Total: Glasgow coma scale total\label{fig.3}} 
\end{figure}

During the experiments, the heat map was employed for model interpretation. We observe that all 12 temporal features contribute significantly to the prediction task, exerting clinical influence on the readmission of essential hypertension patients. The average temporal correlation coefficient is 0.07485428, with six attributes - glucose, heart rate, mean blood pressure, systolic blood pressure, Glasgow Coma Scale eye score, and Glasgow Coma Scale total score - exceeding this mean value.

\section*{Discussion}
  
In the discourse, it becomes evident that deep learning methodologies exhibit superior performance compared to traditional machine learning approaches for feature extraction. Notably, the static and multi-variate temporal fusion network emerges as the top performer among the methods assessed. However, a critical analysis reveals that the feature fusion method STC introduces redundancies by repetitively incorporating static information, potentially hindering effective model training. The  inherent invariance of static data may lead to the generation of noisy data, impeding optimal model performance. Conversely, emphasizing feature extraction from static data through multiple iterations could overlook the crucial temporal vital signs essential for patient discharge decisions. Experimental findings demonstrate that integrating an attention mechanism at each LSTM output step enhances the learning of temporal feature correlations.

Regarding static and temporal fusion strategies, it is observed that employing a gate mechanism alone proves suboptimal. Conversely, the introduction of a self-attention mechanism following the gating mechanism results in improved AUC performance. The designed TSAF approach strategically incorporates static data only once in the final step to mitigate redundancy issues associated with static data integration.

\section*{Conclusions}\label{Conclusion}
In this study, we introduce the Static and Multi-variate Temporal Fusion Network (SMTAFormer) - a novel approach leveraging multi-head self-attention for predicting readmission among ICU patients. By integrating static and temporal feature representations while acknowledging feature correlations, our model utilizes multi-head self-attention to capture both intra-correlations among multivariate temporal features and inter-correlations between static and multi-variate temporal features.

Through experiments conducted on RRA - an essential hypertension readmission dataset extracted from the MIMIC-III database, we achieved promising results with an accuracy of up to 0.866 and an AUC of up to 0.717. Moving forward, our future endeavors include expanding the feature set for enhanced readmission prediction and adapting our methodology to local readmission datasets. Additionally, we aim to address the optimization of patient discharge criteria as a key focus area for further research.

\section*{Declarations}
%%\begin{backmatter}

\section*{Acknowledgments}
%%The authors would like to thank anonymous reviewers for their valuable feedback.

\section*{Funding}%% if any
This work was supported in part by National Science and Technology Major Project under Grant No.2021ZD0111000, in part by the Science and Technology Project of Henan Province under Grant No.232102311232.

%%\section*{Abbreviations}%% if any
%%ACC: accuracy; AUC: area under ROC curve; Bi-LSTM: Bi-directional long short-term Memory; CA: discrete temporal feature extraction module; CGRN: gating residual splicing network; CNN: convolutional neural networks; CO: continuous temporal feature extraction module; DBP: diastolic blood pressure; DNN: fully connected layer; EHR: electronic health records; Eye: glascow coma scale eye opening; F: static-temporal multimodal feature fusion module; GRU: gated recurrent unit; HR: heart rate; ICD: international classification of diseases; LR: linear regression; LSTM: long-short-term memory; MBP: mean blood pressure; MCN: multiple scale convolution; MCNL: multiple scale convolution learnable addition; Motor: glascow coma scale motor response; NB: naive bayes; OS: oxygen saturation; Pre: precision rate;  Rec: recall rate;  ReLU: rectified linear unit; RR: respiratory rate; RF: random forest; RNN: recurrent neural networks; ROC: the receiver operating characteristic; S: static features extraction module; SA: self-attention; SAF: directly self-attention mechanism fusion; SBP: systolic blood pressure; STC: the static features are repeatedly spliced into the model; SVM: support vector machine; T: temporal features extraction module (CO \& CA); Total: glascow coma scale total; TSAF: the self-attention is used twice for static and temporal fusion; Verbal: glascow coma scale verbal response.

\section*{Availability of data and materials}%% if any
The MIMIC-III dataset is available at https://mimic.physionet.org/.

\section*{Authors' contributions}
ZS and RL proposed the method and designed the experiments; ZS and JW performed the experiments; GC and HZ gave advice on methods; SY and LM gave medical advice; ZS and RL wrote the article. All authors read and approved the final manuscript.

\section*{Competing interests}
The authors declare that they have no competing interests.

\section*{Ethics approval and consent to participate}%% if any
The establishment of this database was approved by the Massachusetts Institute of Technology (Cambridge, MA) and Beth Israel Deaconess Medical Center (Boston, MA), and consent was obtained for the original data collection. Data are publicly available (in the MIMIC-III database), so an ethical approval statement and informed consent are not required for the study.

%%\section*{Consent for publication}%% if any
%%Not applicable.

%%\end{backmatter}
%%----end the blinded submitting
\bibliography{sn-article}% common bib file


\begin{thebibliography}{31}
\ifx \bisbn   \undefined \def \bisbn  #1{ISBN #1}\fi
\ifx \binits  \undefined \def \binits#1{#1}\fi
\ifx \bauthor  \undefined \def \bauthor#1{#1}\fi
\ifx \batitle  \undefined \def \batitle#1{#1}\fi
\ifx \bjtitle  \undefined \def \bjtitle#1{#1}\fi
\ifx \bvolume  \undefined \def \bvolume#1{\textbf{#1}}\fi
\ifx \byear  \undefined \def \byear#1{#1}\fi
\ifx \bissue  \undefined \def \bissue#1{#1}\fi
\ifx \bfpage  \undefined \def \bfpage#1{#1}\fi
\ifx \blpage  \undefined \def \blpage #1{#1}\fi
\ifx \burl  \undefined \def \burl#1{\textsf{#1}}\fi
\ifx \doiurl  \undefined \def \doiurl#1{\url{https://doi.org/#1}}\fi
\ifx \betal  \undefined \def \betal{\textit{et al.}}\fi
\ifx \binstitute  \undefined \def \binstitute#1{#1}\fi
\ifx \binstitutionaled  \undefined \def \binstitutionaled#1{#1}\fi
\ifx \bctitle  \undefined \def \bctitle#1{#1}\fi
\ifx \beditor  \undefined \def \beditor#1{#1}\fi
\ifx \bpublisher  \undefined \def \bpublisher#1{#1}\fi
\ifx \bbtitle  \undefined \def \bbtitle#1{#1}\fi
\ifx \bedition  \undefined \def \bedition#1{#1}\fi
\ifx \bseriesno  \undefined \def \bseriesno#1{#1}\fi
\ifx \blocation  \undefined \def \blocation#1{#1}\fi
\ifx \bsertitle  \undefined \def \bsertitle#1{#1}\fi
\ifx \bsnm \undefined \def \bsnm#1{#1}\fi
\ifx \bsuffix \undefined \def \bsuffix#1{#1}\fi
\ifx \bparticle \undefined \def \bparticle#1{#1}\fi
\ifx \barticle \undefined \def \barticle#1{#1}\fi
\bibcommenthead
\ifx \bconfdate \undefined \def \bconfdate #1{#1}\fi
\ifx \botherref \undefined \def \botherref #1{#1}\fi
\ifx \url \undefined \def \url#1{\textsf{#1}}\fi
\ifx \bchapter \undefined \def \bchapter#1{#1}\fi
\ifx \bbook \undefined \def \bbook#1{#1}\fi
\ifx \bcomment \undefined \def \bcomment#1{#1}\fi
\ifx \oauthor \undefined \def \oauthor#1{#1}\fi
\ifx \citeauthoryear \undefined \def \citeauthoryear#1{#1}\fi
\ifx \endbibitem  \undefined \def \endbibitem {}\fi
\ifx \bconflocation  \undefined \def \bconflocation#1{#1}\fi
\ifx \arxivurl  \undefined \def \arxivurl#1{\textsf{#1}}\fi
\csname PreBibitemsHook\endcsname

%%% 1
\bibitem[\protect\citeauthoryear{Rosenberg and Watts}{2000}]{b1}
\begin{barticle}
\bauthor{\bsnm{Rosenberg}, \binits{A.L.}},
\bauthor{\bsnm{Watts}, \binits{C.}}:
\batitle{Patients readmitted to icus: a systematic review of risk factors and outcomes}.
\bjtitle{Chest}
\bvolume{118}(\bissue{2}),
\bfpage{492}--\blpage{502}
(\byear{2000})
\end{barticle}
\endbibitem

%%% 2
\bibitem[\protect\citeauthoryear{Rosenberg et~al.}{2001}]{b2}
\begin{barticle}
\bauthor{\bsnm{Rosenberg}, \binits{A.L.}},
\bauthor{\bsnm{Hofer}, \binits{T.P.}},
\bauthor{\bsnm{Hayward}, \binits{R.A.}},
\bauthor{\bsnm{Strachan}, \binits{C.}},
\bauthor{\bsnm{Watts}, \binits{C.M.}}:
\batitle{Who bounces back? physiologic and other predictors of intensive care unit readmission}.
\bjtitle{Critical care medicine}
\bvolume{29}(\bissue{3}),
\bfpage{511}--\blpage{518}
(\byear{2001})
\end{barticle}
\endbibitem

%%% 3
\bibitem[\protect\citeauthoryear{Frost et~al.}{2009}]{b3}
\begin{barticle}
\bauthor{\bsnm{Frost}, \binits{S.A.}},
\bauthor{\bsnm{Alexandrou}, \binits{E.}},
\bauthor{\bsnm{Bogdanovski}, \binits{T.}},
\bauthor{\bsnm{Salamonson}, \binits{Y.}},
\bauthor{\bsnm{Davidson}, \binits{P.M.}},
\bauthor{\bsnm{Parr}, \binits{M.J.}},
\bauthor{\bsnm{Hillman}, \binits{K.M.}}:
\batitle{Severity of illness and risk of readmission to intensive care: a meta-analysis}.
\bjtitle{Resuscitation}
\bvolume{80}(\bissue{5}),
\bfpage{505}--\blpage{510}
(\byear{2009})
\end{barticle}
\endbibitem

%%% 4
\bibitem[\protect\citeauthoryear{Kaben et~al.}{2008}]{b4}
\begin{barticle}
\bauthor{\bsnm{Kaben}, \binits{A.}},
\bauthor{\bsnm{Corr{\^e}a}, \binits{F.}},
\bauthor{\bsnm{Reinhart}, \binits{K.}},
\bauthor{\bsnm{Settmacher}, \binits{U.}},
\bauthor{\bsnm{Gummert}, \binits{J.}},
\bauthor{\bsnm{Kalff}, \binits{R.}},
\bauthor{\bsnm{Sakr}, \binits{Y.}}:
\batitle{Readmission to a surgical intensive care unit: incidence, outcome and risk factors}.
\bjtitle{Critical care}
\bvolume{12}(\bissue{5}),
\bfpage{1}--\blpage{12}
(\byear{2008})
\end{barticle}
\endbibitem

%%% 5
\bibitem[\protect\citeauthoryear{Ho et~al.}{2009}]{b5}
\begin{barticle}
\bauthor{\bsnm{Ho}, \binits{K.M.}},
\bauthor{\bsnm{Dobb}, \binits{G.J.}},
\bauthor{\bsnm{Lee}, \binits{K.Y.}},
\bauthor{\bsnm{Finn}, \binits{J.}},
\bauthor{\bsnm{Knuiman}, \binits{M.}},
\bauthor{\bsnm{Webb}, \binits{S.A.}}:
\batitle{The effect of comorbidities on risk of intensive care readmission during the same hospitalization: a linked data cohort study}.
\bjtitle{Journal of critical care}
\bvolume{24}(\bissue{1}),
\bfpage{101}--\blpage{107}
(\byear{2009})
\end{barticle}
\endbibitem

%%% 6
\bibitem[\protect\citeauthoryear{Ponzoni et~al.}{2017}]{b56}
\begin{barticle}
\bauthor{\bsnm{Ponzoni}, \binits{C.R.}},
\bauthor{\bsnm{Corr{\^e}a}, \binits{T.D.}},
\bauthor{\bsnm{Filho}, \binits{R.R.}},
\bauthor{\bsnm{Serpa~Neto}, \binits{A.}},
\bauthor{\bsnm{Assun{\c{c}}{\~a}o}, \binits{M.S.}},
\bauthor{\bsnm{Pardini}, \binits{A.}},
\bauthor{\bsnm{Schettino}, \binits{G.P.}}:
\batitle{Readmission to the intensive care unit: incidence, risk factors, resource use, and outcomes. a retrospective cohort study}.
\bjtitle{Annals of the American Thoracic Society}
\bvolume{14}(\bissue{8}),
\bfpage{1312}--\blpage{1319}
(\byear{2017})
\end{barticle}
\endbibitem

%%% 7
\bibitem[\protect\citeauthoryear{Metnitz et~al.}{2003}]{b6}
\begin{barticle}
\bauthor{\bsnm{Metnitz}, \binits{P.G.}},
\bauthor{\bsnm{Fieux}, \binits{F.}},
\bauthor{\bsnm{Jordan}, \binits{B.}},
\bauthor{\bsnm{Lang}, \binits{T.}},
\bauthor{\bsnm{Moreno}, \binits{R.}},
\bauthor{\bsnm{Le~Gall}, \binits{J.-R.}}:
\batitle{Critically ill patients readmitted to intensive care units—lessons to learn?}
\bjtitle{Intensive care medicine}
\bvolume{29}(\bissue{2}),
\bfpage{241}--\blpage{248}
(\byear{2003})
\end{barticle}
\endbibitem

%%% 8
\bibitem[\protect\citeauthoryear{Brown et~al.}{2013}]{b57}
\begin{barticle}
\bauthor{\bsnm{Brown}, \binits{S.E.}},
\bauthor{\bsnm{Ratcliffe}, \binits{S.J.}},
\bauthor{\bsnm{Halpern}, \binits{S.D.}}:
\batitle{An empirical derivation of the optimal time interval for defining icu readmissions}.
\bjtitle{Medical care}
\bvolume{51}(\bissue{8}),
\bfpage{706}
(\byear{2013})
\end{barticle}
\endbibitem

%%% 9
\bibitem[\protect\citeauthoryear{Hosein et~al.}{2013}]{b58}
\begin{barticle}
\bauthor{\bsnm{Hosein}, \binits{F.S.}},
\bauthor{\bsnm{Bobrovitz}, \binits{N.}},
\bauthor{\bsnm{Berthelot}, \binits{S.}},
\bauthor{\bsnm{Zygun}, \binits{D.}},
\bauthor{\bsnm{Ghali}, \binits{W.A.}},
\bauthor{\bsnm{Stelfox}, \binits{H.T.}}:
\batitle{A systematic review of tools for predicting severe adverse events following patient discharge from intensive care units}.
\bjtitle{Critical Care}
\bvolume{17}(\bissue{3}),
\bfpage{1}--\blpage{10}
(\byear{2013})
\end{barticle}
\endbibitem

%%% 10
\bibitem[\protect\citeauthoryear{Brindise and Steele}{2018}]{2018Machine}
\begin{bchapter}
\bauthor{\bsnm{Brindise}, \binits{L.R.}},
\bauthor{\bsnm{Steele}, \binits{R.J.}}:
\bctitle{Machine learning-based pre-discharge prediction of hospital readmission}.
In: \bbtitle{International Conference on Computer, Information and Telecommunication Systems}
(\byear{2018})
\end{bchapter}
\endbibitem

%%% 11
\bibitem[\protect\citeauthoryear{Chung et~al.}{2014}]{b29}
\begin{botherref}
\oauthor{\bsnm{Chung}, \binits{J.}},
\oauthor{\bsnm{Gulcehre}, \binits{C.}},
\oauthor{\bsnm{Cho}, \binits{K.}},
\oauthor{\bsnm{Bengio}, \binits{Y.}}:
Empirical evaluation of gated recurrent neural networks on sequence modeling.
arXiv preprint arXiv:1412.3555
(2014)
\end{botherref}
\endbibitem

%%% 12
\bibitem[\protect\citeauthoryear{Hochreiter and Schmidhuber}{1997}]{b30}
\begin{barticle}
\bauthor{\bsnm{Hochreiter}, \binits{S.}},
\bauthor{\bsnm{Schmidhuber}, \binits{J.}}:
\batitle{Long short-term memory}.
\bjtitle{Neural computation}
\bvolume{9}(\bissue{8}),
\bfpage{1735}--\blpage{1780}
(\byear{1997})
\end{barticle}
\endbibitem

%%% 13
\bibitem[\protect\citeauthoryear{Graves and Schmidhuber}{2005}]{b32}
\begin{barticle}
\bauthor{\bsnm{Graves}, \binits{A.}},
\bauthor{\bsnm{Schmidhuber}, \binits{J.}}:
\batitle{Framewise phoneme classification with bidirectional lstm and other neural network architectures}.
\bjtitle{Neural networks}
\bvolume{18}(\bissue{5-6}),
\bfpage{602}--\blpage{610}
(\byear{2005})
\end{barticle}
\endbibitem

%%% 14
\bibitem[\protect\citeauthoryear{Lin et~al.}{2019}]{b9}
\begin{barticle}
\bauthor{\bsnm{Lin}, \binits{Y.-W.}},
\bauthor{\bsnm{Zhou}, \binits{Y.}},
\bauthor{\bsnm{Faghri}, \binits{F.}},
\bauthor{\bsnm{Shaw}, \binits{M.J.}},
\bauthor{\bsnm{Campbell}, \binits{R.H.}}:
\batitle{Analysis and prediction of unplanned intensive care unit readmission using recurrent neural networks with long short-term memory}.
\bjtitle{PloS one}
\bvolume{14}(\bissue{7}),
\bfpage{0218942}
(\byear{2019})
\end{barticle}
\endbibitem

%%% 15
\bibitem[\protect\citeauthoryear{Morid et~al.}{2020}]{b10}
\begin{barticle}
\bauthor{\bsnm{Morid}, \binits{M.A.}},
\bauthor{\bsnm{Sheng}, \binits{O.R.L.}},
\bauthor{\bsnm{Kawamoto}, \binits{K.}},
\bauthor{\bsnm{Abdelrahman}, \binits{S.}}:
\batitle{Learning hidden patterns from patient multivariate time series data using convolutional neural networks: A case study of healthcare cost prediction}.
\bjtitle{Journal of Biomedical Informatics}
\bvolume{111},
\bfpage{103565}
(\byear{2020})
\end{barticle}
\endbibitem

%%% 16
\bibitem[\protect\citeauthoryear{Zhang et~al.}{2018}]{b11}
\begin{barticle}
\bauthor{\bsnm{Zhang}, \binits{J.}},
\bauthor{\bsnm{Kowsari}, \binits{K.}},
\bauthor{\bsnm{Harrison}, \binits{J.H.}},
\bauthor{\bsnm{Lobo}, \binits{J.M.}},
\bauthor{\bsnm{Barnes}, \binits{L.E.}}:
\batitle{Patient2vec: A personalized interpretable deep representation of the longitudinal electronic health record}.
\bjtitle{IEEE Access}
\bvolume{6},
\bfpage{65333}--\blpage{65346}
(\byear{2018})
\end{barticle}
\endbibitem

%%% 17
\bibitem[\protect\citeauthoryear{An et~al.}{2019}]{b12}
\begin{barticle}
\bauthor{\bsnm{An}, \binits{Y.}},
\bauthor{\bsnm{Huang}, \binits{N.}},
\bauthor{\bsnm{Chen}, \binits{X.}},
\bauthor{\bsnm{Wu}, \binits{F.}},
\bauthor{\bsnm{Wang}, \binits{J.}}:
\batitle{High-risk prediction of cardiovascular diseases via attention-based deep neural networks}.
\bjtitle{IEEE/ACM transactions on computational biology and bioinformatics}
\bvolume{18}(\bissue{3}),
\bfpage{1093}--\blpage{1105}
(\byear{2019})
\end{barticle}
\endbibitem

%%% 18
\bibitem[\protect\citeauthoryear{Choi et~al.}{2016}]{b13}
\begin{botherref}
\oauthor{\bsnm{Choi}, \binits{E.}},
\oauthor{\bsnm{Bahadori}, \binits{M.T.}},
\oauthor{\bsnm{Schuetz}, \binits{A.}},
\oauthor{\bsnm{Stewart}, \binits{W.F.}},
\oauthor{\bsnm{Sun}, \binits{J.}}:
Retain: Interpretable predictive model in healthcare using reverse time attention mechanism. corr abs/1608.05745 (2016).
arXiv preprint arXiv:1608.05745
(2016)
\end{botherref}
\endbibitem

%%% 19
\bibitem[\protect\citeauthoryear{Ma et~al.}{2017}]{b14}
\begin{bchapter}
\bauthor{\bsnm{Ma}, \binits{F.}},
\bauthor{\bsnm{Chitta}, \binits{R.}},
\bauthor{\bsnm{Zhou}, \binits{J.}},
\bauthor{\bsnm{You}, \binits{Q.}},
\bauthor{\bsnm{Sun}, \binits{T.}},
\bauthor{\bsnm{Gao}, \binits{J.}}:
\bctitle{Dipole: Diagnosis prediction in healthcare via attention-based bidirectional recurrent neural networks}.
In: \bbtitle{Proceedings of the 23rd ACM SIGKDD International Conference on Knowledge Discovery and Data Mining},
pp. \bfpage{1903}--\blpage{1911}
(\byear{2017})
\end{bchapter}
\endbibitem

%%% 20
\bibitem[\protect\citeauthoryear{Ma et~al.}{2020}]{b16}
\begin{bchapter}
\bauthor{\bsnm{Ma}, \binits{L.}},
\bauthor{\bsnm{Zhang}, \binits{C.}},
\bauthor{\bsnm{Wang}, \binits{Y.}},
\bauthor{\bsnm{Ruan}, \binits{W.}},
\bauthor{\bsnm{Wang}, \binits{J.}},
\bauthor{\bsnm{Tang}, \binits{W.}},
\bauthor{\bsnm{Ma}, \binits{X.}},
\bauthor{\bsnm{Gao}, \binits{X.}},
\bauthor{\bsnm{Gao}, \binits{J.}}:
\bctitle{Concare: Personalized clinical feature embedding via capturing the healthcare context}.
In: \bbtitle{Proceedings of the AAAI Conference on Artificial Intelligence},
vol. \bseriesno{34},
pp. \bfpage{833}--\blpage{840}
(\byear{2020})
\end{bchapter}
\endbibitem

%%% 21
\bibitem[\protect\citeauthoryear{Vaswani et~al.}{2017}]{b17}
\begin{botherref}
\oauthor{\bsnm{Vaswani}, \binits{A.}},
\oauthor{\bsnm{Shazeer}, \binits{N.}},
\oauthor{\bsnm{Parmar}, \binits{N.}},
\oauthor{\bsnm{Uszkoreit}, \binits{J.}},
\oauthor{\bsnm{Jones}, \binits{L.}},
\oauthor{\bsnm{Gomez}, \binits{A.N.}},
\oauthor{\bsnm{Kaiser}, \binits{{\L}.}},
\oauthor{\bsnm{Polosukhin}, \binits{I.}}:
Attention is all you need.
Advances in neural information processing systems
\textbf{30}
(2017)
\end{botherref}
\endbibitem

%%% 22
\bibitem[\protect\citeauthoryear{Gerrard et~al.}{2022}]{b18}
\begin{bchapter}
\bauthor{\bsnm{Gerrard}, \binits{L.}},
\bauthor{\bsnm{Peng}, \binits{X.}},
\bauthor{\bsnm{Clarke}, \binits{A.}},
\bauthor{\bsnm{Schlegel}, \binits{C.}},
\bauthor{\bsnm{Jiang}, \binits{J.}}:
\bctitle{Predicting outcomes for cancer patients with transformer-based multi-task learning}.
In: \bbtitle{Australasian Joint Conference on Artificial Intelligence},
pp. \bfpage{381}--\blpage{392}
(\byear{2022}).
\bcomment{Springer}
\end{bchapter}
\endbibitem

%%% 23
\bibitem[\protect\citeauthoryear{Luo et~al.}{2020}]{b19}
\begin{bchapter}
\bauthor{\bsnm{Luo}, \binits{J.}},
\bauthor{\bsnm{Ye}, \binits{M.}},
\bauthor{\bsnm{Xiao}, \binits{C.}},
\bauthor{\bsnm{Ma}, \binits{F.}}:
\bctitle{Hitanet: Hierarchical time-aware attention networks for risk prediction on electronic health records}.
In: \bbtitle{Proceedings of the 26th ACM SIGKDD International Conference on Knowledge Discovery \& Data Mining},
pp. \bfpage{647}--\blpage{656}
(\byear{2020})
\end{bchapter}
\endbibitem

%%% 24
\bibitem[\protect\citeauthoryear{Zhang et~al.}{2020}]{b20}
\begin{bchapter}
\bauthor{\bsnm{Zhang}, \binits{X.}},
\bauthor{\bsnm{Qian}, \binits{B.}},
\bauthor{\bsnm{Cao}, \binits{S.}},
\bauthor{\bsnm{Li}, \binits{Y.}},
\bauthor{\bsnm{Chen}, \binits{H.}},
\bauthor{\bsnm{Zheng}, \binits{Y.}},
\bauthor{\bsnm{Davidson}, \binits{I.}}:
\bctitle{Inprem: an interpretable and trustworthy predictive model for healthcare}.
In: \bbtitle{Proceedings of the 26th ACM SIGKDD International Conference on Knowledge Discovery \& Data Mining},
pp. \bfpage{450}--\blpage{460}
(\byear{2020})
\end{bchapter}
\endbibitem

%%% 25
\bibitem[\protect\citeauthoryear{Li et~al.}{2020}]{b21}
\begin{barticle}
\bauthor{\bsnm{Li}, \binits{Y.}},
\bauthor{\bsnm{Rao}, \binits{S.}},
\bauthor{\bsnm{Solares}, \binits{J.R.A.}},
\bauthor{\bsnm{Hassaine}, \binits{A.}},
\bauthor{\bsnm{Ramakrishnan}, \binits{R.}},
\bauthor{\bsnm{Canoy}, \binits{D.}},
\bauthor{\bsnm{Zhu}, \binits{Y.}},
\bauthor{\bsnm{Rahimi}, \binits{K.}},
\bauthor{\bsnm{Salimi-Khorshidi}, \binits{G.}}:
\batitle{Behrt: transformer for electronic health records}.
\bjtitle{Scientific reports}
\bvolume{10}(\bissue{1}),
\bfpage{1}--\blpage{12}
(\byear{2020})
\end{barticle}
\endbibitem

%%% 26
\bibitem[\protect\citeauthoryear{Lin et~al.}{2018}]{b22}
\begin{bchapter}
\bauthor{\bsnm{Lin}, \binits{C.}},
\bauthor{\bsnm{Zhang}, \binits{Y.}},
\bauthor{\bsnm{Ivy}, \binits{J.}},
\bauthor{\bsnm{Capan}, \binits{M.}},
\bauthor{\bsnm{Arnold}, \binits{R.}},
\bauthor{\bsnm{Huddleston}, \binits{J.M.}},
\bauthor{\bsnm{Chi}, \binits{M.}}:
\bctitle{Early diagnosis and prediction of sepsis shock by combining static and dynamic information using convolutional-lstm}.
In: \bbtitle{2018 IEEE International Conference on Healthcare Informatics (ICHI)},
pp. \bfpage{219}--\blpage{228}
(\byear{2018}).
\bcomment{IEEE}
\end{bchapter}
\endbibitem

%%% 27
\bibitem[\protect\citeauthoryear{Li et~al.}{2021}]{b23}
\begin{bchapter}
\bauthor{\bsnm{Li}, \binits{D.}},
\bauthor{\bsnm{Lyons}, \binits{P.G.}},
\bauthor{\bsnm{Klaus}, \binits{J.}},
\bauthor{\bsnm{Gage}, \binits{B.F.}},
\bauthor{\bsnm{Kollef}, \binits{M.}},
\bauthor{\bsnm{Lu}, \binits{C.}}:
\bctitle{Integrating static and time-series data in deep recurrent models for oncology early warning systems.}
In: \bbtitle{CIKM},
pp. \bfpage{913}--\blpage{936}
(\byear{2021})
\end{bchapter}
\endbibitem

%%% 28
\bibitem[\protect\citeauthoryear{Lim et~al.}{2021}]{b24}
\begin{barticle}
\bauthor{\bsnm{Lim}, \binits{B.}},
\bauthor{\bsnm{Ar{\i}k}, \binits{S.{\"O}.}},
\bauthor{\bsnm{Loeff}, \binits{N.}},
\bauthor{\bsnm{Pfister}, \binits{T.}}:
\batitle{Temporal fusion transformers for interpretable multi-horizon time series forecasting}.
\bjtitle{International Journal of Forecasting}
\bvolume{37}(\bissue{4}),
\bfpage{1748}--\blpage{1764}
(\byear{2021})
\end{barticle}
\endbibitem

%%% 29
\bibitem[\protect\citeauthoryear{An et~al.}{2021}]{b25}
\begin{bchapter}
\bauthor{\bsnm{An}, \binits{Y.}},
\bauthor{\bsnm{Zhang}, \binits{H.}},
\bauthor{\bsnm{Sheng}, \binits{Y.}},
\bauthor{\bsnm{Wang}, \binits{J.}},
\bauthor{\bsnm{Chen}, \binits{X.}}:
\bctitle{Main: Multimodal attention-based fusion networks for diagnosis prediction}.
In: \bbtitle{2021 IEEE International Conference on Bioinformatics and Biomedicine (BIBM)},
pp. \bfpage{809}--\blpage{816}
(\byear{2021}).
\bcomment{IEEE}
\end{bchapter}
\endbibitem

%%% 30
\bibitem[\protect\citeauthoryear{Zhang et~al.}{2020}]{b26}
\begin{bchapter}
\bauthor{\bsnm{Zhang}, \binits{X.}},
\bauthor{\bsnm{Xiao}, \binits{C.}},
\bauthor{\bsnm{Glass}, \binits{L.M.}},
\bauthor{\bsnm{Sun}, \binits{J.}}:
\bctitle{Deepenroll: patient-trial matching with deep embedding and entailment prediction}.
In: \bbtitle{Proceedings of The Web Conference 2020},
pp. \bfpage{1029}--\blpage{1037}
(\byear{2020})
\end{bchapter}
\endbibitem

%%% 31
\bibitem[\protect\citeauthoryear{Johnson et~al.}{2016}]{b8}
\begin{barticle}
\bauthor{\bsnm{Johnson}, \binits{A.E.}},
\bauthor{\bsnm{Pollard}, \binits{T.J.}},
\bauthor{\bsnm{Shen}, \binits{L.}},
\bauthor{\bsnm{Lehman}, \binits{L.-w.H.}},
\bauthor{\bsnm{Feng}, \binits{M.}},
\bauthor{\bsnm{Ghassemi}, \binits{M.}},
\bauthor{\bsnm{Moody}, \binits{B.}},
\bauthor{\bsnm{Szolovits}, \binits{P.}},
\bauthor{\bsnm{Anthony~Celi}, \binits{L.}},
\bauthor{\bsnm{Mark}, \binits{R.G.}}:
\batitle{Mimic-iii, a freely accessible critical care database}.
\bjtitle{Scientific data}
\bvolume{3}(\bissue{1}),
\bfpage{1}--\blpage{9}
(\byear{2016})
\end{barticle}
\endbibitem

\end{thebibliography}

\end{document}